\pdfoutput=1

\documentclass[11pt]{article}

\usepackage[]{acl}

\usepackage{times}
\usepackage{latexsym}
\usepackage{url}
\usepackage{paralist}
\usepackage{multirow}
\usepackage{verbatim}
\usepackage{amsfonts}
\usepackage{amssymb}
\usepackage{amsmath}
\usepackage{algorithm}
\usepackage{algpseudocode}
\usepackage{graphics}
\usepackage{epsfig}
\usepackage{multirow}
\usepackage{makecell}
\usepackage{booktabs}
\usepackage[scientific-notation=true]{siunitx}

\usepackage{tabularx}
\newcolumntype{Y}{>{\centering\arraybackslash}X}

\newcommand{\StatexIndent}[1][3]{%
  \setlength\@tempdima{\algorithmicindent}%
  \Statex\hskip\dimexpr#1\@tempdima\relax}

\newcolumntype{R}[1]{>{\raggedleft\let\newline\\\arraybackslash\hspace{0pt}}m{#1}}

\usepackage[T1]{fontenc}

\usepackage[utf8]{inputenc}

\usepackage{microtype}

\title{BARCOR: Towards A Unified Framework for\\ Conversational Recommendation Systems}

\author{Ting-Chun Wang\quad Shang-Yu Su\quad Yun-Nung Chen\\
National Taiwan University, Taipei, Taiwan \\
  \texttt{\{b06901061,f05921117\}@ntu.edu.tw\quad y.v.chen@ieee.org} \\}

\begin{document}
\maketitle
\begin{abstract}
Recommendation systems focus on helping users find items of interest in the situations of information overload, where users' preferences are typically estimated by the past observed behaviors. 
In contrast, conversational recommendation systems (CRS) aim to understand users' preferences via interactions in conversation flows.
CRS is a complex problem that consists of two main tasks: (1) recommendation and (2) response generation.
Previous work often tried to solve the problem in a modular manner, where recommenders and response generators are separate neural models. 
Such modular architectures often come with a complicated and unintuitive connection between the modules, leading to inefficient learning and other issues.
In this work, we propose a unified framework based on BART for conversational recommendation, which tackles two tasks in a single model.
Furthermore, we also design and collect a lightweight knowledge graph for CRS in the movie domain.
The experimental results show that the proposed methods achieve the state-of-the-art performance in terms of both automatic and human evaluation.
\footnote{The data and source code will be released once accepted.}
\end{abstract}

\section{Introduction}
Though recommendation systems have gained tremendous success in various domains and many aspects of our lives, they have potential limitations.
Practically, recommending is often a one-shot, reactive, uni-directional process. Users passively receive recommended information from the systems in certain pre-defined situations.
It assumes that a user has clear, immediate requests when interacting with the system; however, such recommending may not be accurate since user demand would change over time and vacillate.
Sometimes users are indecisive; to this end, traditional recommendation systems lack proactive guidance.
Conversational Recommendation Systems (CRS) became an emerging research topic, focusing on exploring users' preferences through natural language interaction.
Generally speaking, CRSs support goal-oriented, multi-turn dialogues, which proactively acquire precise user demand by interactions.
Thereby, CRS is a complex system consisting of a recommendation module and a dialogue module, which make suitable recommendations and generate proper responses respectively.

In terms of modeling, CRS requires seamless integration between the recommendation module and the dialogue module.
The systems need to understand user preferences by preceding dialogue context and recommend suitable items. To recommend items to users in the natural language form, the generated responses need to contain relevant items while being fluent and grammatically correct. 
Previous work has proposed different approaches for integrating the two major modules, for instance, building belief trackers over semi-structured user queries \cite{sun2018conversational, Zhang_2020} and switching decoders for component selection \cite{li2018deep}.
Furthermore, as practical goal-oriented dialogue systems, CRSs usually utilize Knowledge Graphs (KG) for introducing external knowledge and system scalability.
Choosing a suitable KG, leveraging the information of entities, and interacting with the two main components of CRS for high-quality recommendation is undoubtedly another challenging problem.

Recent work \cite{zhou2020improving} proposed to incorporate two special KGs for enhancing data representations of both components and fuse the two semantic spaces by associating two different KGs. 
Specifically, they incorporate ConceptNet \cite{speer2017conceptnet} for word-level information and DBpedia \cite{lehmann2015dbpedia} for item information.
ConceptNet provides word information such as synonyms and antonyms of certain words, which helps understand dialogue context. At the same time, DBpedia has structural information of entities, providing rich attributes and direct relations between items.
However, these public large-scale knowledge graphs were not designed for CRSs hence may not be suitable.
Though prior methods have achieved some improvement in performance, there are some potential limitations.
Most of them build recommender and response generator separately with complicated and unintuitive connection between the modules, which may cause inefficient learning and unclear knowledge transfer between the modules.
For example, the work mentioned above \cite{zhou2020improving} requires training multiple graph convolution networks for KG embeddings, mutual information maximization to bridge the embedding spaces.  
In this case, the practical usage and scalability of the system design are a concern to some extent.

To this end, we propose a unified framework for the conversational recommendation, which tackles two tasks in a single model.
The framework is built on top of pretrained BART \cite{Lewis_2020} and finetuned on the recommendation and response generation tasks.
We proposed to use the bidirectional encoder of BART as the recommender and the auto-regresive decoder as the response generator, so-called \textbf{BARCOR} (\textbf{B}idirectional \textbf{A}uto-\textbf{R}egressive \textbf{CO}nversational \textbf{R}ecommender).
Moreover, we design and collect a lightweight knowledge graph for CRS in the movie domain.
With the essentially-connected model structure of BART, we do not need to worry about designing a connection between the recommender and the response generator.

To sum up, the contributions can be summarized as 3-fold:
\begin{compactitem}
\item This paper proposes a general framework conversational recommendation based on BART, which tackles two tasks in a single model.
\item This work designs and collects a lightweight knowledge graph for CRS in the movie domain.
\item The benchmark experiments demonstrate the effectiveness of the proposed framework.
\end{compactitem}

\begin{figure*}[t!]
\centering 
\includegraphics[width=0.9\linewidth]{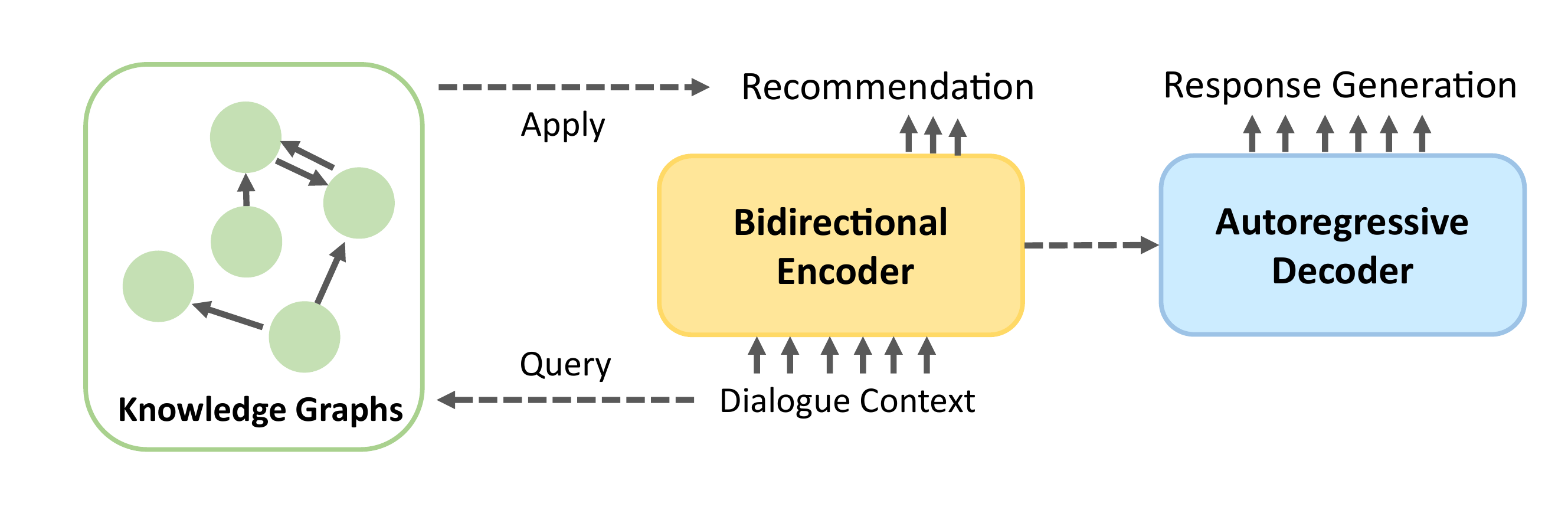}
 \vspace{-7mm}
\caption{The proposed framework is composed of three components: (1) knowledge graphs for providing external knowledge, (2) a bidirectional encoder as the recommender, and (3) an auto-regressive decoder as the response generator.} 
\label{fig:framework} 
\end{figure*}

\section{Related Work}


As a specific type of goal-oriented dialogue systems, Conversational Recommendation Systems (CRS) have also moved towards the use of neural networks \cite{li2018deep}.
\citet{christakopoulou2018q} uses recurrent neural network-based models to recommend videos to users; \citet{zhang2016collaborative} explores the use of knowledge bases in recommendation tasks.
\citet{sun2018recurrent} proposes an embedding-based approach to learn semantic representations of entities and paths in a KG to characterize user preferences towards items.
\citet{wang2019kgat} improves the performance of the recommenders by learning the embeddings for entities in the KG using the TransR algorithm \cite{lin2015learning} and refining and discriminating the node embeddings by using attention over the neighbour nodes of a given node. 
\citet{wang2018ripplenet} and \citet{li2020seamlessly} focus on solving the task of goal-oriented conversation recommendation for cold-start users.
\citet{li2020seamlessly} generates new venues for recommendation using graph convolution networks (GCNs) and encodes the dialogue contents using hierarchical recurrent encoder-decoder (HRED) \cite{sordoni2015hierarchical} and thereby recommend locations to users.
\citet{li2018deep} released the ReDial dataset wherein users are recommended movies based on the conversation they have with the recommendation agents. 
KBRD \cite{chen2019towards} extends the work of \citet{li2018deep} by incorporating a KG and proposing a graph-based recommender for movie recommendations.
They have also shown that dialogue and recommendation in CRSs are complementary tasks and benefit one another. 
To better understand user's preferences, KGSF \cite{zhou2020improving} introduces a word-oriented KG to facilitate node representation learning. 
Recently, to generate natural and informative responses with accurate recommendations, \citet{lu-etal-2021-revcore} incorporates movie reviews, and \citet{zhang2021kecrs} proposes supervision signals for the semantic fusion of words and entities.


\section{Dataset}
\label{sec:dataset}
The ReDial \cite{li2018deep} dataset is widely adopted for the conversational recommendation task. This dataset is constructed through Amazon Mechanical Turk (AMT) and comprises multi-turn conversations centered around movie recommendations in seeker-recommender pairs. It contains 10,006 conversations consisting of 182,150 utterances related to 51,699 movies.

To generate training data, previous work \cite{zhou2020improving} viewed all items mentioned by recommenders as recommendations. 
However, this processing measure causes issues, clearly stated in \citet{zhang2021kecrs}. 
First, repetitive items are likely to guide a model to simply recommend items once appeared in dialogues.
Secondly, the evaluation dataset is biased to repetitive recommendations, failing to present recommendation quality faithfully.
To address the issues, we only consider items as recommendations only if they aren't mentioned before.

Since the recommendation module takes over the item recommendation task, the dialogue module could focus on capturing sentence semantics to produce fluent conversations.
Thus, we mask the recommended items in the target response with a special token, $\texttt{[MOVIE]}$. 
It also serves as a placeholder for items retrieved by the recommender module in generated responses during the inference phase. Table \ref{tab:filtering-methods} shows training examples from this process.

\begin{table*}[t!]
    \centering
    \resizebox{1.0\textwidth}{!}{
        \begin{tabular}{ | c | c | l | l | c | }
            \hline
            \multicolumn{1}{|c|}{} & 
            \multicolumn{1}{c|}{\bf Accepted} & 
            \multicolumn{1}{c|}{\bf Context} &
            \multicolumn{1}{c|}{\bf Response} & 
            \multicolumn{1}{c|}{\bf Target movie} \\
        
            \hline \hline
            \multirow{3}{*}{(a)} & \multirow{3}{*}{\includegraphics[height=0.02\textheight]{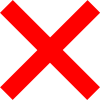}} & S: Hi, I am looking for a movie like Super Troopers.  & \multirow{3}{*}{Yes $\texttt{[MOVIE]}$ is funny.}  & \multirow{3}{*}{\textbf{Police Academy}} \\
            {} & {} & R: You should watch \textbf{Police Academy}. & {} & {}  \\
            {} & {} & S: Is that a great one? I have never seen it. & {} & {}  \\
            
            \hline
            \multirow{4}{*}{(b)} & \multirow{4}{*}{\includegraphics[height=0.02\textheight]{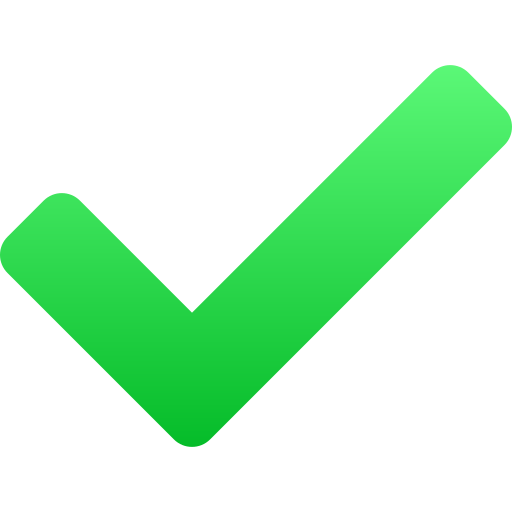}} & R: Hello, what kind of movies do you like? & {} & \multirow{4}{*}{\textbf{Happy Death Day}}  \\ 
            {} & {} & {S: I am looking for a movie recommendation.} & {Oh, you like scary movies?} & {}  \\
            {} & {} & {S: When I was younger, I really enjoyed the} & {I recently watched $\texttt{[MOVIE]}$. } & {}  \\
            {} & {} & {  A Nightmare on Elm Street.}  & {} & {}  \\
        
            \hline
        \end{tabular}
    }
    \vspace{-1mm}
    \caption{Examples in the processed ReDial dataset. In the column of context, "S" and "R" represent a movie seeker and a recommender respectively. Recommended items in responses are masked by $\texttt{[MOVIE]}$. Example (a) isn't accepted to the processed dataset since "Police Academy" is a repetitive item, which is presented in the context.}
    \vspace{-2mm}
    \label{tab:filtering-methods}
\end{table*}

\section{Preliminaries}
In this section, we first introduce the problem formulation and then detail the collected knowledge graph.

\subsection{Problem Formulation}
For the dataset, $\{u_i\}^{n}$ denotes a conversation, where $u_i$ is the utterance at $i$-th turn, and $n$ is the number of conversation history. 
We process a conversation into multiple data triplets $(X, \mathcal{I}, y)$. 
At $j$-th turn, $X_j=\{u_i\}^{j-1}_{i=1}$ denotes the conversation context, $\mathcal{I}_j$ is the set of ground truth items presented in $u_j$ for the recommendation task, and $y_j=u_j$ denotes the target response for the generation task. 
Note that every entry in $\mathcal{I}_j$ cannot appear in the context $X_j$ as stated in the previous section, and it can be an empty set when there is no need for recommendations. 
For the knowledge graph, $\mathcal{G}=\{(e_h, r, e_t)|e_h, e_t \in \mathcal{E}, r\in \mathcal{R}\}$ denotes the KG, where $(e_h, r, e_t)$ means the head entity $e_h$ and the tail entity $e_t$ is related by the relation $r$. 
The entity set $\mathcal{E}$ consists of a movie item set $\mathcal{I}$ and a set of descriptive entities that are film properties. 
The set of ground truth items $\mathcal{I}_j$ is the subset of $\mathcal{I}$.

The conversational recommendation is essentially the combination of two tasks: document retrieval and natural language generation. 
They are formulated as two objective functions, $f(X, \mathcal{G})$ and $g(X, \mathcal{I}_{\text{pred}})$. 
$f(X, \mathcal{G})$ gives novel recommendations  $\mathcal{I}_{\text{pred}}$ based on the context $X$ and the KG $\mathcal{G}$, and $g(X, \mathcal{I}_{\text{pred}})$ generates natural responses based on the context and the recommended items.

\subsection{CORG (COnversational Recommender Graphs)}
In the previous work, a wide variety of external knowledge sources are incorporated to facilitate recommendations. 
However, the KGs adopted in the previous work \cite{zhou2020improving,chen2019towards, sarkar2020kgrec} are open-domain KGs, e.g., DBpedia and ConceptNet, which may introduce too many irrelevant entities and obscure high-order connectivity as stated in \citet{zhang2021kecrs}.
Although some datasets, MindReader \cite{2020MindReader} is intended for movie recommendations, its coverage of movies in the ReDial dataset is low, as shown in Table \ref{tab:KG-characteristics}.
To mitigate these issues, we construct a knowledge graph called \textbf{CORG} (\textbf{CO}nversational \textbf{R}ecommender \textbf{G}raphs), which contains 5 types of node entities and 5 types of relations.

\paragraph{Data Source}
We collect information of movies from Wikidata\footnote{\url{https://www.wikidata.org/wiki/Wikidata:Main_Page}}, which is a collaboratively edited multilingual knowledge graph hosted by Wikimedia Foundation\footnote{\url{https://wikimediafoundation.org/}}.
It contains movie-related information and identifiers of other databases for additional information, such as synopses or reviews.

\paragraph{Information Collection}
Nodes in CORG comprise two kinds of entities: \emph{movies items} and \emph{descriptive entities}.
Movies items are all mentioned movies in ReDial, and descriptive entities are associative properties of those movies.
We use  "movie name" and "release year" as keys to query Wikidata to collect movie properties, including movie genres, cast members, directors, and production companies.
In this way, we get the entire set of nodes in CORG, whose statistics are shown in Table \ref{tab:corg-stats}.
Among 6,924 mentioned movies in ReDial, CORG covers 6,905 movies (99.7\%).

\paragraph{Data Processing}
Assuming seekers are only interested in protagonists, we select top-10 main cast members.
Besides, since movie genres in Wikidata are hierarchically arranged (e.g, superhero film is a subclass of action and adventure films), we recursively build edges between the nodes of genres and those of their parent genres.
The edge statistics are shown in Table \ref{tab:corg-stats}.

\begin{table*}[t!]
    \centering
    \resizebox{1.0\textwidth}{!}{   
        \begin{tabular}{ | l | c | c | c | c | }
            \hline
            
            \bf Knowledge Graph &  \bf \# Movies &   \bf \# Entities &  \bf Designed for ReDial &  \bf Movie Coverage for ReDial \\\hline \hline
            
            MinderReader \cite{2020MindReader} & 4,941 & 18,707 & \includegraphics[height=0.01\textheight]{figures/icons/x.png} & 44.6\% \\ \hline
            
            
            DBpedia (KGSF) \cite{zhou2020improving}  & 6,111 & 64,361 & \includegraphics[height=0.01\textheight]{figures/icons/check.png} & 88.2\% \\ \hline
            
            TMDKG \cite{zhang2021kecrs}  & 6,692 & 15,822 & \includegraphics[height=0.01\textheight]{figures/icons/check.png} & 96.2\% \\ \hline
            
            \textbf{CORG} & \textbf{6,905} & \textbf{23,164} & \includegraphics[height=0.01\textheight]{figures/icons/check.png} & \textbf{99.7}\% \\ \hline
            
        \end{tabular}
    }
    \caption{Characteristics of CORG and existing knowledge graphs. Although TMDKG has high movie coverage, their source code is not publicly available.}
    \label{tab:KG-characteristics}
\end{table*}

        

\begin{figure}[t!]
    \centering 
    \includegraphics[width=\linewidth]{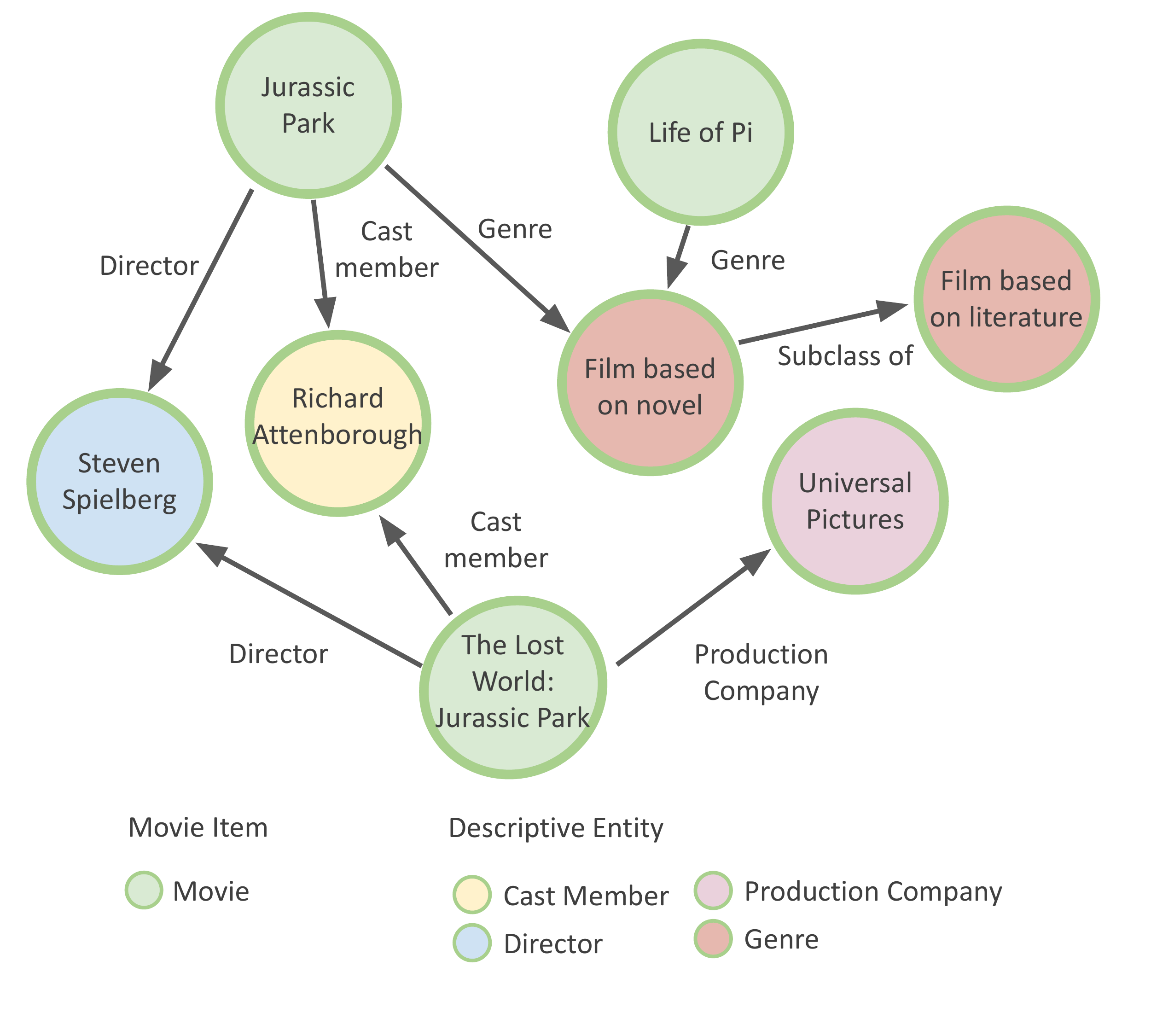}
    \vspace{-10mm}
    \caption{A sample subgraph of CORG. CORG has 5 types of node entities and 5 types of relations, the statistics of types and relations are shown in Table \ref{tab:corg-stats}.}
    \label{fig:sample-corg} 
\end{figure}

\section{BARCOR}
\label{sec:barcor}

We propose to use the bidirectional encoder of BART \cite{Lewis_2020} as the recommender and the auto-regresive decoder as the response generator, so-called \textbf{BARCOR} (\textbf{B}idirectional \textbf{A}uto-\textbf{R}egressive \textbf{CO}nversational \textbf{R}ecommender).
BARCOR is a unified framework for the conversational recommendation which tackles two tasks in a single model.
The proposed framework is composed of three main components: (1) a knowledge graph encoder to provide external knowledge, (2) a bidirectional encoder for recommendation, and (3) an auto-regressive decoder for response generation.
In this section, we will go through the design of each component in the pipeline.

\subsection{Graph Encoder}
\label{ssec:graph_encoder}
We follow \citet{zhou2020improving}, adopting Relational Graph Convolutional Network (R-GCN) \cite{schlichtkrull2017modeling} to learn entity representations in CORG.
Formally, the hidden state of an entity $i$ at the ($l+1$)-th layer is formulated as:
\begin{equation*}
\mathbf{h}_i^{(l+1)}=\sigma(\sum_{r\in\mathcal{R}}\sum_{j\in\mathcal{E}_i^r}\dfrac{1}{|\mathcal{E}_i^r|}\mathbf{W}_r^{(l)}\mathbf{h}_j^{(l)}+\mathbf{W}^{(l)}\mathbf{h}_i^{(l)}),
\end{equation*}
where $\mathbf{h}_i^{(l)} \in \mathbb{R}^{d_E}$ is the hidden state of the entity $i$ at the $l$-th layer, $d_E$ is the dimension of the hidden state, and $\mathbf{h}_i^0$ is also referred to as the entity embedding $\mathbf{e}_i$.
$\mathcal{E}_i^r$ is the set of neighboring entities of the entity $i$ related by $r$, and its cardinality serves as a normalization constant.
$\mathbf{W}_r^{(l)}$ denotes a learnable relation-specific transformation matrix for the hidden states of neighboring entities under a relation $r$, and $\mathbf{W}^{(l)}$ is a learnable matrix for transforming hidden states at the $l$-th layer.
We treat the hidden states of the last layer as the representations of entities in CORG, which is denoted by $\mathbf{H}\in\mathbb{R}^{(|\mathcal{E}|\times d_E)}$.
The representations construct a search space of recommended candidates for item retrieval.

Other than the recommendation task, we include the node classification task to facilitate graph representation learning. Given an entity representation $\mathbf{h}$ and a multiple layer perceptron (MLP), we obtain a node type prediction $\mathbf{p}_{\text{node}} \in \mathbb{R}^{N_T}$, where $N_T$ is the number of node types:
\begin{equation}
    \mathbf{p}_{\text{node}} = \mathrm{Softmax}(\mathrm{MLP}(\mathbf{h})).
    \label{eq:node-loss}
\end{equation}
Then, we conduct a cross entropy loss $L_{\text{node}}$ between the prediction from Equation (\ref{eq:node-loss}) and ground truth node types to optimize the graph encoder.

\subsection{BART as Conversational Recommender}
\label{ssec:bart}
BART is a Transformer-based \cite{vaswani2017attention} sequence-to-sequence model, which can be seen as the generalizing BERT (bidirecitonal encoder) and GPT (autoregressive decoder).
In the design of BART, the decoder performs cross-attention from each of its layers over the final hidden state of the encoder to be aware of input sequences.
This operation seamlessly integrates the recommendation and dialogue modules into a unified conversational recommender. 

BARCOR features four advantages over the graph-based recommender in the previous works: 
First, a unified framework inherently fuses the semantics between the encoder and the decoder and becomes less sensitive to the design of model architecture and hyper-parameters selections. 
In contrast, other works propose complex attentive interactions between modules, which is not robust from an actual production system perspective.
That is, slight parameter changes would impact the performance.
Moreover, BART is proved to be effective in various downstream tasks, such as neural machine translation and question answering.
Secondly, BART takes users' utterances as inputs without further processing.
Instead, in \citet{zhou2020improving}, the graph-based recommender demands manual annotations for movies and words in input texts to build a user preference, which is impractical under a realistic scenario.
Thirdly, the learned knowledge from pretrained models provides rich sentence semantics.
Finally, BART can perform an end-to-end training scheme for both the recommendation and generation tasks.
Conversely, other works tend to design separate modules for two tasks and further sequentially optimize each module.

\paragraph{Bidirectional Recommender}
Given a conversation context $X$, BART encoder transforms $X$ into $\mathbf{c}$, the hidden state of the final self-attentive layer.
Then, $\mathbf{c}$ is viewed as a sentence representation of $X$ and also a search key for retrieving recommendation candidates.
To derive the probability over the candidates, we apply inner-product to compute the similarity between $\mathbf{c}$ and entity representations $\mathbf{H}$ from the graph encoder,
\begin{align}
        \mathbf{p}_{\text{rec}}=\mathrm{Softmax}(\mathbf{c}\mathbf{H}^{\intercal}),  \label{eq:p_rec} \\
        \mathbf{p}_{\text{rec-infer}}=\mathrm{Softmax}(\mathbf{c}\mathbf{H}_I^{\intercal}), \label{eq:p_rec_inf}
\end{align}
where $\mathbf{p}_{\text{rec}} \in \mathbb{R}^{|\mathcal{E}|}$ denotes the recommendation prediction.
To learn parameters in BARCOR, we employ a cross-entropy loss $L_{\text{rec}}$ between the prediction from Equation (\ref{eq:p_rec}) and the labels of ground truth entities.
Note that the search space of recommended candidates is $\mathbf{H}$, which means both \emph{movie items} and \emph{descriptive entities} are likely to be retrieved. 

\paragraph{Data Augmentation}
Since sentence-level semantics extracted from BART encoder is naturally inconsistent with entity-level semantics from the graph encoder, other than optimizing BARCOR by $L_{\text{rec}}$, we propose to (1) augment the training set with descriptive entities and (2) strategically initialize the graph encoder's embeddings to facilitate the fusion of heterogeneous semantics. 
First, during training, we construct data using the names of descriptive entities as the conversation context, such as "George Clooney," and the entities themselves as the recommended items.
The data allows the representations of descriptive entities to be directly optimized by  $L_{\text{rec}}$ instead of optimized indirectly through their one-hop neighboring movie items.
Besides, BARCOR becomes more aware of their names in conversation context and neighboring movie items.
Secondly, we initialize entity embeddings $\{\mathbf{e}_i\}_{i=1}^{|\mathcal{E}|}$ with the sentence representations of their names transformed by the pretrained BART encoder.
Thus, the initial semantic gap between two types of representations becomes closer,  presumably easier to fuse.
However, during the inference phase, the search space is reduced to the item set $\mathcal{I}$. The recommendation prediction is computed through Equation (\ref{eq:p_rec_inf}), where $\mathbf{H}_{I}$ is the matrix only consisting of movie item representations.

\paragraph{Auto-Regressive Response Generator}
We retain the original operations of BART decoder, which is conditioned on an input sequence and its sentence representation (i.e., the final hidden state of BART encoder) to generate a response auto-regressively.
Therefore, we follow \citet{Radford2018ImprovingLU} to compute the generative probability and optimize the decoder through negative log-likelihood.
During training, we mask the target responses of the augmented dataset to preserve authentic conversation flows.

\paragraph{End-to-End Training}
We optimize BARCOR by simultaneously performing the recommendation and generation tasks, compared to previous works demanding sequential optimization for two separated components. 
That is, we jointly minimize the objective as follow:
\begin{equation*}
    L=L_{\text{rec}} + \alpha L_{\text{gen}} + \beta L_{\text{node}},
\end{equation*}
where $\alpha$ and $\beta$ are hyper-parameters determined by cross-validation.

\section{Experiments}
\subsection{Experiment Setup}
\paragraph{Baselines}
We compare BARCOR with the following baseline methods for the recommendation and response generation tasks on the processed ReDial dataset as discussed in Section \ref{sec:dataset}.
\begin{compactitem}
    \item \textbf{KBRD} \cite{chen2019towards} employs DBpedia to enhance semantics of contextual items or entities for the construction of user preferences. The dialogue module is based on Transformer, where KG information is incorporated as word bias during generation.
    \item \textbf{KGSF} \cite{zhou2020improving} uses MIM \cite{MIM} to fuse the information of entity-oriented and word-oriented KGs (i.e., DBpedia and ConceptNet). A user preference is constructed by fused representations of items and words. The dialogue module is based on Transformer, consisting of a standard encoder and a KG-enhanced decoder.
\end{compactitem}

\paragraph{Automatic Evaluation}
For the recommendation task, we adopt \emph{Recall@k} (R@k, k=1, 5, 10, 50), which suggests whether top-k recommended items contain the ground truth recommendations for evaluation.
Since users may be frustrated by too many recommendations within a response, Recall@1,5 more faithfully present the recommendation performance.
For the generation task, we follow \citet{zhou2020improving} to use \emph{Distinct n-gram (Dist-n, n=2, 3, 4)}, which measures the diversity of sentences.
Since CRSs interact with humans through natural language, we introduce two metrics to capture the effectiveness of recommendations.
\emph{Item-F1} measures whether a CRS accurately provides recommendations compared to ground truth responses. 
\emph{Average Item Number (AIN)} denotes the average number of recommended items within a sentence and presents the informativeness of generated responses.

\paragraph{Human Evaluation}
Aligning the CRS goal of providing successful recommendations, we invite 11 professional annotators to judge response quality. 
Given 40 multi-turn conversations from the testing set, the annotators evaluate the quality in terms of three aspects: (1) \emph{Fluency}, (2) \emph{Relevancy}, and (3) \emph{Informativeness}, with each score ranging from 0 to 2.


\subsection{Result Analysis}
\begin{table*}[t!]
\centering
    \resizebox{1.0\textwidth}{!}{
    \begin{tabular}{ | c | l | c c c c | c c c c c | c c c| }
        \hline
        \multicolumn{2}{|c|}{\multirow{2}{*}{\bf Model}} & \multicolumn{4}{c|}{\multirow{1}{*}{\bf Recommendation}} & \multicolumn{5}{c|}{\bf Response Generation}   & \multicolumn{3}{c|}{\bf Human Evaluation} \\
        \multicolumn{2}{|c|}{} & 
        \bf R@1 & \bf R@5 & \bf R@10 & \bf R@50 & 
        \bf Dist-2  & \bf Dist-3 & \bf Dist-4 & \bf Item-F1 & \bf AIN &
        \bf Fluen. & \bf Relev. & \bf Informat. \\ 
        \hline \hline
        (a) & KBRD & 1.46 & 7.23 & 12.65 & 30.26 & 14.32  & 27.27 & 39.57 & 58.80 & 36.63 & 1.62 & 1.08 & 1.01 \\
        (b) & KGSF & 1.41 & 7.66 & 13.47 & 32.17 & 19.49  & 35.36 & 49.19 & 62.61 & 41.00 & 1.56 & 0.98 & 0.66 \\
        (c) & \bf BARCOR & \bf 2.53 & \bf 9.98 & \bf 16.17 & \bf 34.95 & \bf 58.90  & \bf 88.75 & \bf 102.52 & \bf 71.71 & \bf 53.00 & \bf 1.86 & \bf 1.76 & \bf 1.57 \\
        \hline
        (e) & (c) - Node Loss & 2.32 & 9.01 & 15.61 & 34.3 & 41.12  & 61.15 & 73.60 & 71.08 & 45.22 & - & - & - \\
        (f) & (c) - Data Aug. & 2.23 & 9.22 & 14.62 & 34.16 & 31.91  & 45.05 & 53.57 & 55.13 & 44.64 & - & - & - \\
        (g) & (c) - Node Init. & 1.95 & 8.68 & 14.67 & 33.86 & 22.32  & 35.33 & 45.19 & 68.21 & 44.30 & - & - & - \\
        (h) & (c) - CORG & 2.29 & 9.15 & 15.32 & 33.34 & 30.50  & 43.11 & 50.80 & 70.00 & 48.37 & - & - & - \\
        \hline
    \end{tabular}
    }
    \vspace{-1mm}
    \caption{Results on the recommendation and response generation tasks. In human evaluation, ``Fluen.'', ``Relev'', and ``Informat'' denote fluency, relevancy, and informativeness, respectively. The best results are in bold.}
    \vspace{-2mm}
    \label{tab:results}
\end{table*}

Table \ref{tab:results} summarizes the performance of different methods on the ReDial dataset, including human evaluation and automatic evaluation for the recommendation and response generation tasks.

\paragraph{Item Recommendation}
As we can see, KGSF outperforms KBRD because KGSF incorporates a word-oriented KG to enrich entity representations, highlighting the importance of words in context for the representation learning.
With learned knowledge from pretrained models, BARCOR achieves 2.53\% in R@1, 9.98\% in R@5, 16.17\% in R@10, and 34.95\% in R@50 and outperforms KGSF by 79\% and 30\% in terms of R@1 and R@5 respectively.
It demonstrates a tight fusion of semantics between sentences in context and entities in KG.
Also, context and knowledge provide richer entity information, compared to the word-oriented KG adopted by KGSF. 

\paragraph{Response Generation}
In the automatic evaluation, the proposed BARCOR outperforms all baseline methods with a large margin in terms of Dist-n.
Compared to KGSF, it improves Dist-2, Dist-3, and Dist-4 by +39.41\%, +53.39\%, and +53.33\%, respectively, which demonstrates the proposed method effectively generates diverse sentences.
Besides, BARCOR achieves 71.71\% in Item-F1 and 53\% in AIN. It suggests that BARCOR interprets user intentions to further precisely generate responses containing recommendations. 
In the human evaluation, BARCOR performs best among all baseline methods for the three metrics.
We can note that BARCOR especially has higher scores of Relevancy and Informativeness, indicating generated responses are both accurately aligned with user intentions and rich in recommended items and related information.
It verifies our interpretation of the scores of Item-F1 and AIN in the automatic evaluation.
The above results prove the effectiveness of our method that fuses entity representations from the KG with sentence representations to generate fluent, relevant, and informative utterances.
We also provide qualitative analysis in Appendix \ref{appendix:qual-analysis}.

\paragraph{Training Stability}
\label{system-stability}
\begin{figure}[t!]
    \centering 
    \includegraphics[width=\linewidth]{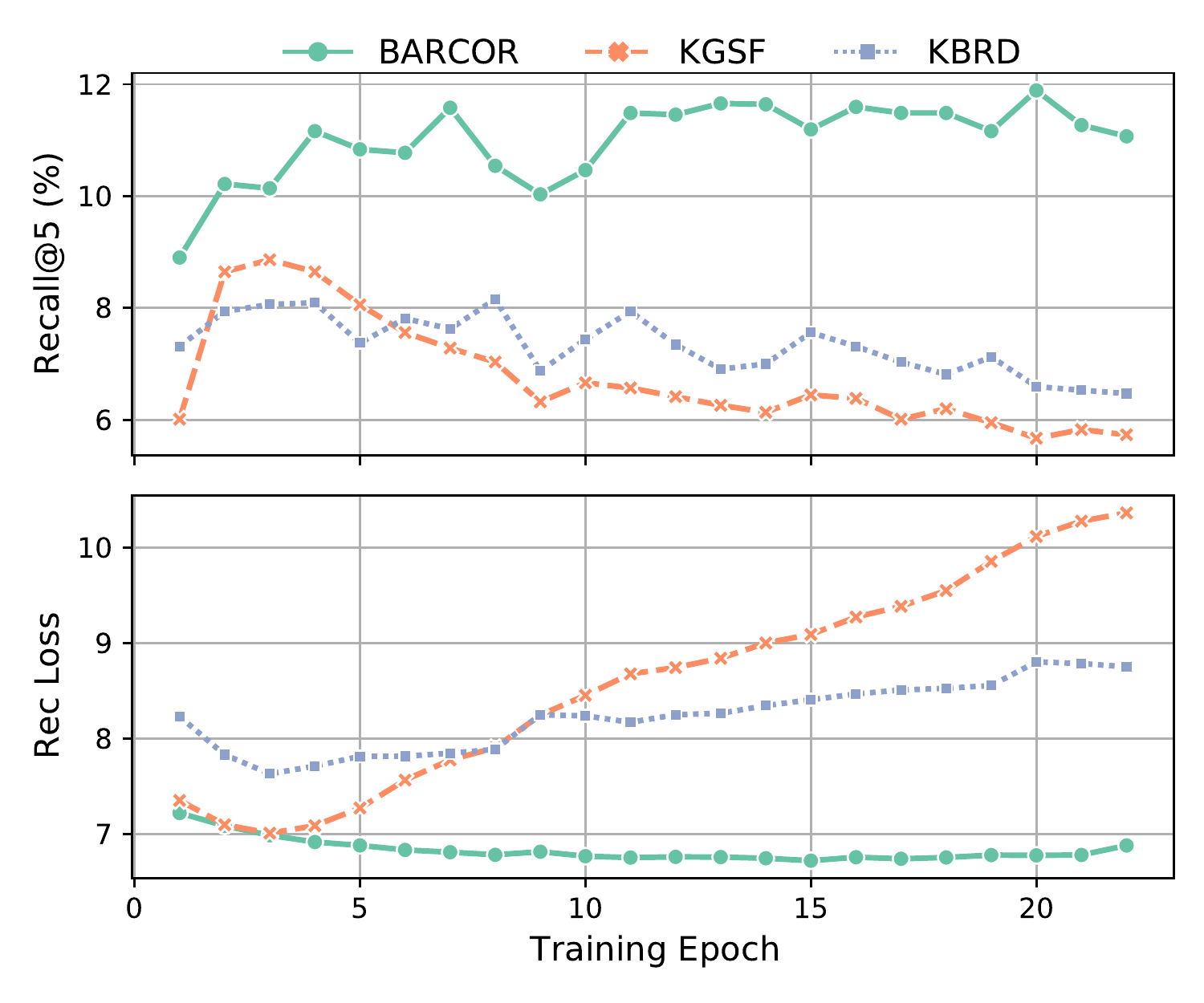}
    \vspace{-6mm}
    \caption{Recommendation performance of BARCOR and the baselines on the validation set at different training epochs.}
    \vspace{-4mm}
    \label{fig:model_stability}
\end{figure}

Figure \ref{fig:model_stability} shows the performance curves of Recall@5 (R@5) and recommendation loss on the validation set for different methods.
We select R@5 as the evaluation metric since it is neither too strict nor tolerable for accurate recommendations. It can be observed that BARCOR is more stably optimized and achieves a better performance than other competitive baseline methods.
Within the first four epochs, both KBRD and KGSF quickly reach an optimal state where models gain the highest R@5 with the least recommendation loss. 
However, as training progresses, they begin to overfit the training data, leading to the decline in R@5 and the rise of the recommendation loss. 
The instability may be attributed to the insufficiency of semantics in conversation context and the number of trainable parameters. 
To construct a user representation, the baselines aggregate information of annotated entities, including movies and their associative properties, in conversation context. 
Although KGSF incorporates a word-oriented KG and a semantic fusion technique, the combinations of words and entities are still limited to the training set and the KGs. 
Therefore, some informative words or entities and their variants are lost if not presented in the corpus. 
In contrast, BARCOR directly encodes an entire context to build a user representation, ensuring every word is considered and increasing word semantic richness. 
Learned knowledge from pretrained models also prevents BARCOR from overly biasing on the training set. 
Moreover, we note that the number of trainable parameters of the BARCOR's recommendation module (39 million) is less than half of that of KGSF's (106 million) and KBRD's (91 million) recommenders.
More details about models is presented in Table \ref{tab:model-stats} in Appendix. 
Optimized fewer parameters with inputs of richer semantics, BARCOR consistently outperforms these baselines for all recommendation metrics. 
The results demonstrate the effectiveness and optimization stability of the proposed unified framework for modeling CRS.

\subsection{Ablation Study}
\label{ssec:ablation}
\begin{figure}[t!]
    \centering 
    \includegraphics[trim=0.5cm 0cm 0.5cm 0cm, clip=true,width=\linewidth]{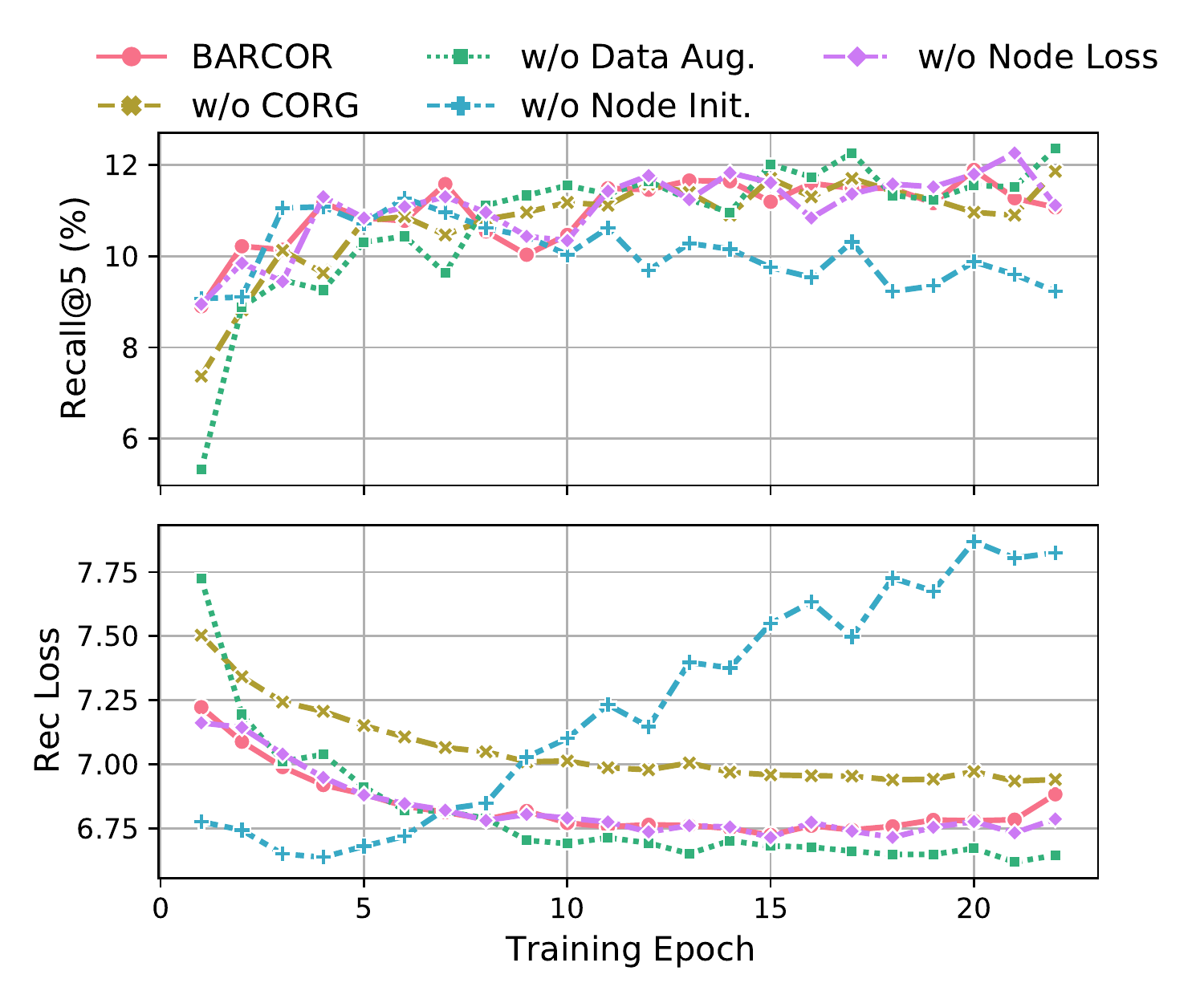}
    \vspace{-6mm}
    \caption{Ablation study: Recommendation performance on the validation set at different training epochs.} 
    \vspace{-4mm}
    \label{fig:ablation}
\end{figure}

To understand the contribution of each component on the recommendation and generation tasks, we construct a ablation study for four variants in BARCOR: 
(1) BARCOR (w/o Node Loss): removing cross entropy loss of the node classification task presented in Section \ref{ssec:graph_encoder},
(2) BARCOR  (w/o Data Aug.): removing the training set augmentation mentioned in Section \ref{ssec:bart},
(3) BARCOR (w/o Node Init.): replacing node embeddings from the pretrained BART encoder by randomly initialized weights mentioned in Section \ref{ssec:bart},
and (4) BARCOR  (w/o CORG): excluding CORG by removing relations among nodes. 

Since the recommendation and dialogue modules share the same sentence representation of context, techniques designed for representation enrichment are mutually beneficial for both tasks. 
As shown in Table \ref{tab:results} (row(e-h)), all techniques are helpful to improve the final performance in terms of all metrics. 
Besides, node embeddings initialization of the graph encoder and the proposed CORG are seemed to be more critical. 
First, we observe that R@1, R@5, and Dist-n decrease when the node embeddings are randomly initialized. 
Also, the validation performance curves in Figure \ref{fig:ablation} reveal the issue of overfitting, as shown in Section \ref{system-stability}. 
We attribute this to the increased optimization difficulty brought by the incorporation of the graph encoder. 
The number of its trainable parameters is 27 million, accounting for 68\% of the total trainable parameters in the recommendation module. 
Randomly initialized embeddings easily fit the seen data but difficultly fuse with sentence semantics from the BARCOR's encoder.
The results reinforce our claim discussed in Section \ref{system-stability}.
Although random initialization leads to the decline in performance, BARCOR (w/o Node Init.) still outperforms the strong baselines for all evaluation metrics. 
Second, as shown in row(h), BARCOR (w/o CORG) surprisingly achieves competitive results with BARCOR in R@1, R@5, and R@10 and outperforms KGSF using two KGs. 
Namely, BARCOR (w/o CORG) merely leverages relations of entities and words in the dialogue history to recommend more accurately than the KG-enhanced strong baselines. 
It implies that implicit relations of entities within context have yet been exploited to the fullest.

In conclusion, the sentence-level semantics derived from BARCOR's encoder provide richer information than the entity representations encoded by the R-GCN, and is sufficient for accurate recommendations. 
Besides, a trade-off between KG-based information enrichment and optimization difficulty for a graph encoder needs careful consideration.
In our work, we propose incorporating supervision signal from the node classification task, training set augmentation, and node embeddings initialized by the pretrained BART to reduce the difficulty. 
We hope these results inspire future research.

\section{Conclusion}
In this paper, we proposed a novel unified framework for the conversational recommendation, BARCOR.
BARCOR jointly tackles the recommendation and generation tasks with the shared sentence representation of conversation history. 
It serves as a search key for item retrieval and provides rich fused semantic of sentences and entities for the decoder to generate responses.
Moreover, we enrich the information of entities by constructing a high-quality KG, namely CORG, and incorporating a graph encoder exploiting structural knowledge. 
The experiments results demonstrate that BARCOR achieves better performance on recommendation accuracy and response quality than all competitive baselines and generates informative responses with great fluency and relevancy.

\bibliography{anthology}
\bibliographystyle{acl_natbib}

\appendix

\clearpage
\newpage


\begin{table}[t!]
    \begin{tabular}{ | p{5cm} | R{1.8cm}| }
            \hline
            \bf Measure & \bf Value \\ \hline\hline
            \# Node & 23,164 \\ \hline
            \# Movie & {6,924} \\ 
            \# Genre & {313} \\ 
            \# Cast Member & {11,017} \\ 
            \# Director & {3,587} \\ 
            \# Production Company & {1,323} \\ \hline \hline
            
            \# Edge & 87,212 \\ \hline
            \# Movie-Genre & {19,292} \\
            \# Movie-Cast Member & {53,109} \\ 
            \# Movie-Director & {7,155} \\
            \# Movie-Production Company & 7,407 \\
            \# Genre-Genre & {249} \\ \hline
    \end{tabular}
    \vspace{-0.5em}
    \caption{Graph statistics of the constructed CORG.}
    \label{tab:corg-stats}
    \vspace{0.5em}
\end{table}

\section{Implementation Details}
\label{appendix:implementation}
In all the experiments, we use mini-batch AdamW with learning rate $\num{0.00003}$ as the optimizer and each batch of 64 examples on a single Nvidia Tesla V100. 
The whole training takes $22$ epochs without early stop. 
The entire implementation was based on PyTorch, PyTorch Geometric \cite{fey2019fast}, and HuggingFace transformers\footnote{\url{https://huggingface.co/transformers/}} package.
We finetune the $11$-th attention layer of BART encoder and the $10$-th and $11$-th attention layers of BART decoder for the CRS task. 
The detailed number of trainable parameters are listed in Table \ref{tab:model-stats}.

\begin{table}[t!]
    \begin{tabular}{ | p{1.5cm} | p{1.3cm} | p{1.3cm} | p{1.8cm} | }
            \hline    
            \bf Model & \bf Rec.  & \bf Gen. & \bf \# Total \\ \hline\hline
            KBRD   & 85.9 \% & 14.1 \% & 105,601,166 \\ \hline  
            KGSF   & 81.6 \% & 18.4 \% & 129,899,342 \\ \hline  
            BARCOR & 53.8 \% & 46.2 \% & 72,593,777 \\ \hline  
    \end{tabular}
    \vspace{-0.5em}
    \caption{Model statistics. "Rec." and "Gen." represent the recommendation and response generation modules, respectively. These two columns present the ratio of the trainable parameters in the two modules to the total number of trainable parameters, which is presented in the column "\# Total." The statistics of KBRD\footnote{\url{https://github.com/THUDM/KBRD}} and KGSF\footnote{\url{https://github.com/Lancelot39/KGSF}} are derived from the publicly released implementations of their authors on GitHub.}
    \label{tab:model-stats}
    \vspace{0.5em}
\end{table}

\section{Qualitative Analysis}
\label{appendix:qual-analysis}
\begin{table}[t!]
    \begin{tabular}{|  p{2cm} p{4.7cm}  |}
        \hline
        \small \bf (a) & {} \\ \hline
        \small \bf Seeker: & \small Hello, I really need a good movie for tonight. I have a date.\\
        \small \bf Recommender: & Ok. \\ 
        \small \bf Seeker: & \small Maybe a romantic comedy like \textbf{How to Lose a Guy in 10 Days (2003)}. \\ \hline
        \small \bf KBRD: & \small I haven't seen that one yet. \\\hline
        \small \bf KGSF: & \small \textbf{Love Actually} is a good one.  \\ \hline
        \small \bf BARCOR: & \small \textbf{50 First Dates (2004)} is a good romantic comedy with Adam Sandler.  \\ \hline
        \small \bf Human: & \small How about \textbf{10 Things I Hate About You (1999)}? \\ \hline\hline
        
        \small \bf (b) & {} \\ \hline
        \small \bf Seeker: & \small Hello! How are you?  \\
        \small \bf Recommender: & \small Hey there. I'm doing well. Can I recommend a movie for you? \\
        \small \bf Seeker: & \small I'm looking for a great old movie. Any suggestions? \\ \hline
        \small \bf KBRD: & \small Sure! Have you seen \textbf{Black Panther (2018)}? \\\hline
        \small \bf KGSF: & \small I am doing great.  \\ \hline
        \small \bf BARCOR: & \small Have you seen \textbf{Gone with the Wind (1939)}?  \\ \hline
        \small \bf Human: & \small Hmmmm, such as \textbf{Breakfast at Tiffany's (1961)}? I loved that movie. \\ \hline\hline
        
        \small \bf (c) & {} \\ \hline
        \small \bf Recommender: & \small Looking for a good movie?  \\
        \small \bf Seeker: & \small I have a movie night tonight. My firends are coming. I need good mystery movies. \\
        \small \bf Recommender: & \small Well, \textbf{Wind River (2017)} is a good mystery.  \\
        \small \bf Seeker: & \small I have seen that one. \\\hline
        \small \bf KBRD: & \small It is really good.\\ \hline
        \small \bf KGSF: & \small \textbf{Shutter Island (2010)} is a good one.  \\ \hline
        \small \bf BARCOR: & \small  \textbf{Gone Girl (2014)} is another good one. If you're looking for something a little more graphic, it is one of my favorite movies.  \\ \hline
        \small \bf Human: & \small How about \textbf{Memento (2000)}? \\ \hline
        
    \end{tabular}

    \caption{Examples of generated responses from different models. Movie names are in bold.}
    \label{tab:examples}
\end{table}

In the section, we present several conversations generated by different models in Table \ref{tab:examples}. 
Compared to other baseline methods, BARCOR can understand the user intention to provide a relevant recommendation and generate informative responses related to the recommended item. 
In example (a), when the seeker asks for romantic comedy and mentions "\emph{How to Lose a Guy in 10 Days (2003)}", BARCOR recommends another romantic comedy  "\emph{50  First  Dates  (2004)}". 
Besides, it also expresses the attitude toward the recommended item and makes the response more informative by saying that "\emph{is a good romantic comedy with Adam Sandler}." 
In example (b), BARCOR grasps the idea of great old movies and recommends "\emph{Gone with the Wind (1939)}",  an epic historical romance film.
Conversely, KBRD simply recommends a well-known modern movie, which fails to meet the user demand.
In example (c), when asked a mystery movie like "\emph{Wind River (2017)}", the human recommender and KGSF merely give recommendations without personal insight. 
However, BARCOR not only recommends another mystery movie, "\emph{Gone Girl (2014)}", but explains the motivation behind the recommendation by saying that "\emph{If you’re looking for something a little more graphic, it is one of my favorite movies}."

\end{document}